\documentclass[
]{ceurart-arxiv}

\usepackage{tcolorbox}
\usepackage{listings}
\begin{document}

\copyrightyear{}
\copyrightclause{}
\conference{}

\title{An Explainable Natural Language Framework for Identifying and Notifying Target Audiences In Enterprise Communication}


\author[1,2]{V\'{i}tor N. Louren\c{c}o}[%
orcid=0000-0003-0711-1339,
email=vitornl@amazon.co.uk
]\cormark[1]

\author[1]{Mohnish Dubey}[%
email=dubeymns@amazon.co.uk
]

\author[3]{Yunfei Bai}[%
orcid=0009-0009-0270-6123,
email=byunfei@amazon.com
]

\author[4]{Audrey Depeige}[%
orcid=0000-0002-7469-2551,
email=audreycd@amazon.lu
]

\author[4]{Vivek Jain}[%
orcid=0009-0009-1743-7923,
email=vivejai@amazon.lu
]

\address[1]{Amazon, London, United Kingdom}
\address[2]{Fluminense Federal University, Rio de Janeiro, Brazil}
\address[3]{Amazon, Seattle, United States}
\address[4]{Amazon, Luxembourg, Luxembourg}

\cortext[1]{Corresponding author.}

\begin{abstract}
  In large-scale maintenance organizations, identifying subject matter experts and managing communications across complex entities relationships poses significant challenges -- including information overload and longer response times -- that traditional communication approaches fail to address effectively. We propose a novel framework that combines RDF graph databases with LLMs to process natural language queries for precise audience targeting, while providing transparent reasoning through a planning-orchestration architecture. Our solution enables communication owners to formulate intuitive queries combining concepts such as equipment, manufacturers, maintenance engineers, and facilities, delivering explainable results that maintain trust in the system while improving communication efficiency across the organization.
\end{abstract}


\maketitle
\section{Introduction}

Amazon's Reliability \& Maintenance Engineering~(RME) is a global organization responsible for maintaining and troubleshooting Material Handling Equipment across Amazon's network of facilities.
As part of its daily operations, RME ensures operational continuity through reactive, preventive and predictive maintenance, managing an extensive asset infrastructure of millions of equipment configurations and parts supplied by thousands different vendors -- from conveyor belts to robots and fire alarms.

In such a complex ecosystem, two critical industrial challenges emerge. First, pinpointing subject matter experts or specific individuals and sites working with particular equipment, parts, or manufacturers becomes increasingly difficult and fails to scale with rapid organizational growth.
Second, the complexity of these relationships creates significant communication bottlenecks.
Traditional approaches of broadcasting updates organization-wide or using manual database queries leads to either missing key recipients or overwhelming others with irrelevant information.
Adding to these challenges, recent advances in Natural Language~(NL) understanding lead users to expect accurate, precise, and reliable results from NL formulated queries like ``\textit{I want to reach out to all maintenance technicians working with Vendor X's conveyor belts or fire alarms of model FA123 at European sites}''.

The scale and complexity of these challenges calls for an innovative approach to managing maintenance relationships and streamlining organizational communications -- one that can effectively address both current operational needs and emerging user expectations. Large Language Models~(LLMs) and chat-inspired solutions present promising avenues to tackle these challenges. However, when deploying LLM-based solutions in enterprise environments, organizations face a major challenge: the lack of transparency in how models arrive at their conclusions~\cite{Nagadivya2023}. 
In RME context the challenge is pronounced, where decisions directly impact the efficiency and timeliness of operational communications across global network. Without adequate explainability and transparency, users risk compromising both the effectiveness of their communications and their trust in the outputs of the AI-based system.

In this paper, we propose a novel framework that addresses these challenges. Our solution empowers communication owners to find any audience groups relevant to their maintenance issue through simple, natural language queries, while providing transparent reasoning for the retrieved results. Our framework eliminates the need for complex database operations or specialized query languages by integrating RDF graph databases with full-text search capabilities, enabling efficient connections between user terms and known entities, as well as facilitating complex queries across various entity relationships (\textit{e.g.}, equipment, people, and sites). To bridge the gap between natural language and structured queries, we implement dual LLM-driven workflows that transform user inputs into graph queries. Additionally, we adopt a planning-orchestration architecture~\cite{sapkota2025aiagentsvsagentic} that generates clear explanations of the system's reasoning process, ensuring transparency and enabling the user to validate the results.

\section{Methodology}

\begin{figure}[ht!]
    \centering
    \includegraphics[width=\linewidth]{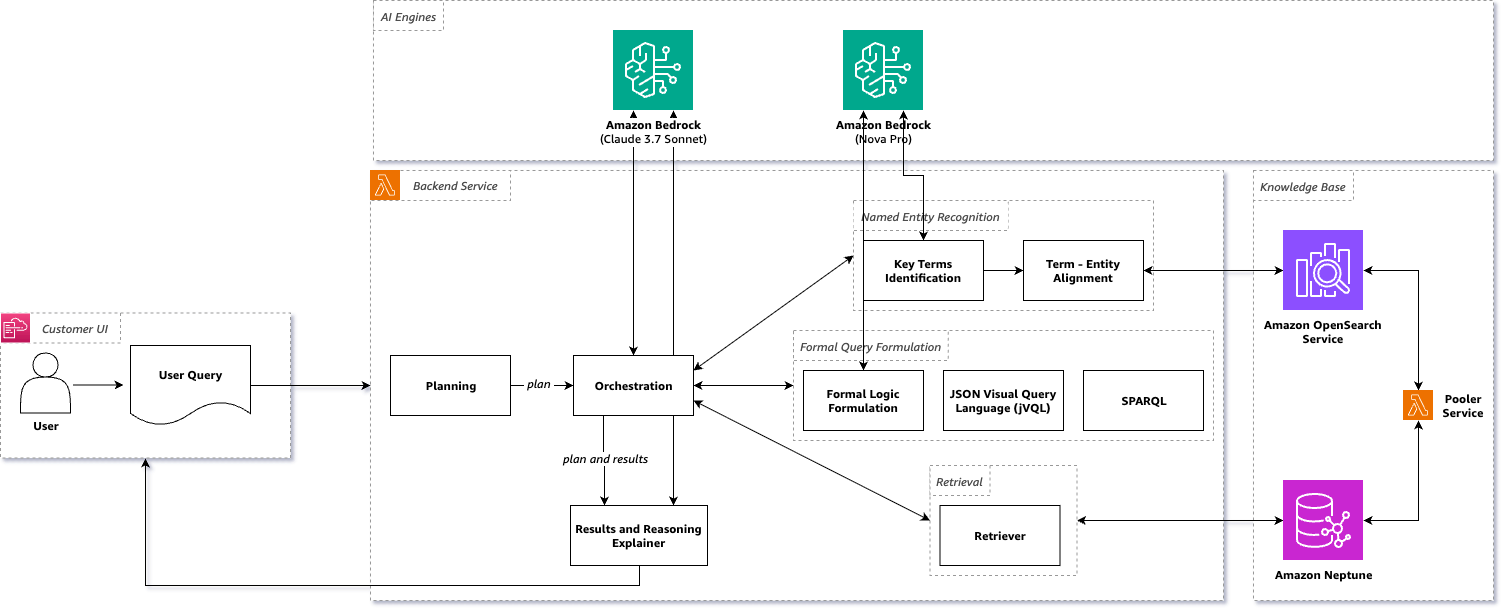}
    \caption{System architecture developed for Amazon Reliability \& Maintenance Engineering.}
    \label{fig:framework}
\end{figure}

The Figure~\ref{fig:framework} illustrates the system's architecture of our solution for Amazon RME is composed of four main components: Customer UI, AI Engines, Knowledge Base, and the proposed framework. We briefly describe the first three components before detailing the proposed framework.

The Customer UI provides three core functionalities: natural language query processing, construction of formal queries, and visualization of results with reasoning explanations. The AI Engines infrastructure layer leverages foundation models available at Amazon Bedrock (Nova Pro~\cite{agi2025amazonnovafamilymodels} and Claude 3.5 Sonnet) for natural language understanding and processing. The Knowledge Base consolidates RME information from multiple internal data sources, including equipment documentation, employee profiles and roles, and facilities data. This information is curated, transformed, and structured according to our purpose-built maintenance ontology. The graph database technology of choice was Amazon Neptune as it supports both RDF graph storage with reasoning capabilities and scalable graph analytics. Further, we integrate the knowledge graph with Amazon's OpenSearch Service to enable efficient entity search using the full-text search functionality, while maintaining the synchronization between the graph database and search index through an event-driven architecture.

\subsection{Proposed Framework}

The proposed framework is the fourth and core component of the system's architecture. The framework starts processing the natural language statement formulated by the communication owner using the planning-orchestration architecture, which acts as an intelligent workflow orchestrator. The goal is to evaluate the users’ query -- to establish an execution plan --, and orchestrating the plan along with the execution workflows (also called \emph{agents}).

Two key aspects of translating Natural Language~(NL) to any query language are interpreting and understanding the key terms -- NL representations of existing entities contained in the NL statement -- and the relationships among them. The proposed framework relies on two architectural components responsible for these tasks. The Named Entity Recognition~(NER) workflow component identifies the key terms and aligns them with their corresponding existing entities. The Formal Query Formulation~(FQF) workflow formally establishes the relationships among these entities as expressed in the NL statement.

For the NER workflow, we adopt a prompt-based approach for identifying key terms using Amazon Nova Pro as the foundation model. The prompt uses few-shot examples and specialized tags to mark the key terms to be processed at a later stage, and requests the model to output the original statement augmented with the marked key terms. This approach incorporates several lessons learned in recognizing named entities using LLMs. We highlight two key findings: 1) constraining the key terms search space helps avoid hallucinations, and 2) using consistent markers ensures output reliability. After obtaining the initial statement with marked key terms, we parse it using a regex expression to extract the terms. For each term, we then proceed to find the best matching entity in the Knowledge Base (\textit{i.e.}, the top-ranked entity label with the highest text matching score).

The FQF workflow follows a similar architecture, using Nova Pro as its based model and few-shot examples prompt to express the natural language statement in a formal logical formulation. Differently from the NER workflow, in this workflow, we use a structured output to retrieve the answer. The output comprises of the formulation of the statement as Boolean algebra expression and a reasoning breakdown of the formulation process. The latter output was a key finding during our research work, where we discovered that a Chain-of-Thought~\cite{Wei2022} inspired reasoning breakdown helps to achieve more accurate Boolean algebra formulations, while helping the reasoning explainer process. With the natural language statement expressed in a Boolean formulation, we augment it by replacing the key terms with their matched entities and associated classes. At this stage, the example user statement ``I want to reach out to all maintenance technicians working with Vendor X’s conveyor belts or fire alarms of model FA123 at European sites'', would have being translated to a formal expression, as:


\begin{footnotesize}
\begin{tcolorbox}[colback=gray!10, boxsep=0pt, left=0pt, right=0pt, frame empty]
\begin{verbatim}
class: <http://info.rme.amazon.dev/ontology/> ; entity: <http://info.rme.amazon.dev/entity/>
class:JobTitle ( entity:MaintenanceTechnician ) AND class:Region ( entity:EU ) AND
( class:Equipment ( entity:ConveyorBelt ) OR class:EquipmentModel ( entity:FA123 ) ) .
\end{verbatim}
\end{tcolorbox}
\end{footnotesize}

This expression is then converted to JSON Visual Query Language (jVQL) -- to visually express the query in the UI -- and into SPARQL -- to query Amazon Neptune and retrieve the corresponding data.

While LLMs can effectively process natural language, their interpretations might not always align with the users' intent. For instance, in a query to ``\textit{find all technicians working with vendor X in LATAM}'', users need to verify that the system correctly identified each entity (\texttt{technicians} as job title, \texttt{X} as vendor, \texttt{LATAM} as region). In RME's complex structure, even simple queries like ``managers at European sites'' require clarity on whether the system includes only managers or also senior leadership positions. Such validation becomes crucial in high-stakes environments where communications must reach the right audience in a timely manner, particularly for operational updates, network actions and safety-related messages.

To address this need for validation, our framework implements an explainability feature based on multi-step reasoning~\cite{paranjape2023artautomaticmultistepreasoning}. The system combines the execution plan steps with their descriptions and intermediate outputs to generate structured, concise explanations. These explanations reveal how the system interprets key terms, maps them to knowledge base entities, and constructs logical relationships for formal queries, enabling users to validate decisions and refine their queries for better accuracy.

\section{Conclusion}

This paper presented a novel framework that addresses audience targeting and communication challenges in large-scale maintenance organizations. By combining RDF graph databases with LLM-driven workflows, our solution enables natural language querying while maintaining transparency through explainable AI techniques. The framework bridges the gap between user intent and formal queries, demonstrating how enterprises can leverage AI capabilities without compromising explainability in critical operational communications. Future work could extend this approach to other domains and explore additional explainability techniques.


\section*{Declaration on Use of Generative AI}
During the preparation of this work, the author(s) used Claude family models and Amazon Nova family models to grammar and spelling check. After using these tool(s)/service(s), the author(s) reviewed and edited the content as needed and take(s) full responsibility for the publication’s content. 

\bibliography{main}

\begin{thebibliography}{5}
\expandafter\ifx\csname natexlab\endcsname\relax\def\natexlab#1{#1}\fi
\providecommand{\url}[1]{\texttt{#1}}
\providecommand{\href}[2]{#2}
\providecommand{\path}[1]{#1}
\providecommand{\DOIprefix}{doi:}
\providecommand{\ArXivprefix}{arXiv:}
\providecommand{\URLprefix}{URL: }
\providecommand{\Pubmedprefix}{pmid:}
\providecommand{\doi}[1]{\href{http://dx.doi.org/#1}{\path{#1}}}
\providecommand{\Pubmed}[1]{\href{pmid:#1}{\path{#1}}}
\providecommand{\bibinfo}[2]{#2}
\ifx\xfnm\relax \def\xfnm[#1]{\unskip,\space#1}\fi
\bibitem[{Balasubramaniam et~al.(2023)Balasubramaniam, Kauppinen, Rannisto, Hiekkanen, and Kujala}]{Nagadivya2023}
\bibinfo{author}{N.~Balasubramaniam}, \bibinfo{author}{M.~Kauppinen}, \bibinfo{author}{A.~Rannisto}, \bibinfo{author}{K.~Hiekkanen}, \bibinfo{author}{S.~Kujala},
\newblock \bibinfo{title}{Transparency and explainability of ai systems: From ethical guidelines to requirements},
\newblock \bibinfo{journal}{Information and Software Technology} \bibinfo{volume}{159} (\bibinfo{year}{2023}) \bibinfo{pages}{107197}. \DOIprefix\doi{https://doi.org/10.1016/j.infsof.2023.107197}.
\bibitem[{Sapkota et~al.(2025)Sapkota, Roumeliotis, and Karkee}]{sapkota2025aiagentsvsagentic}
\bibinfo{author}{R.~Sapkota}, \bibinfo{author}{K.~I. Roumeliotis}, \bibinfo{author}{M.~Karkee},
\newblock \bibinfo{title}{Ai agents vs. agentic ai: A conceptual taxonomy, applications and challenges},
\newblock \bibinfo{journal}{arXiv}  (\bibinfo{year}{2025}). \URLprefix \url{https://arxiv.org/abs/2505.10468}.
\bibitem[{Intelligence(2025)}]{agi2025amazonnovafamilymodels}
\bibinfo{author}{A.~A.~G. Intelligence},
\newblock \bibinfo{title}{The amazon nova family of models: Technical report and model card},
\newblock \bibinfo{journal}{arXiv}  (\bibinfo{year}{2025}). \URLprefix \url{https://arxiv.org/abs/2506.12103}.
\bibitem[{Wei et~al.(2022)Wei, Wang, Schuurmans, Bosma, Ichter, Xia, Chi, Le, and Zhou}]{Wei2022}
\bibinfo{author}{J.~Wei}, \bibinfo{author}{X.~Wang}, \bibinfo{author}{D.~Schuurmans}, \bibinfo{author}{M.~Bosma}, \bibinfo{author}{B.~Ichter}, \bibinfo{author}{F.~Xia}, \bibinfo{author}{E.~H. Chi}, \bibinfo{author}{Q.~V. Le}, \bibinfo{author}{D.~Zhou},
\newblock \bibinfo{title}{Chain-of-thought prompting elicits reasoning in large language models},
\newblock in: \bibinfo{booktitle}{Proceedings of the 36th International Conference on Neural Information Processing Systems}, \bibinfo{year}{2022}.
\bibitem[{Paranjape et~al.(2023)Paranjape, Lundberg, Singh, Hajishirzi, Zettlemoyer, and Ribeiro}]{paranjape2023artautomaticmultistepreasoning}
\bibinfo{author}{B.~Paranjape}, \bibinfo{author}{S.~Lundberg}, \bibinfo{author}{S.~Singh}, \bibinfo{author}{H.~Hajishirzi}, \bibinfo{author}{L.~Zettlemoyer}, \bibinfo{author}{M.~T. Ribeiro}, \bibinfo{title}{Art: Automatic multi-step reasoning and tool-use for large language models}, \bibinfo{year}{2023}. \URLprefix \url{https://arxiv.org/abs/2303.09014}. \href{http://arxiv.org/abs/2303.09014}{{\tt arXiv:2303.09014}}.

\end{thebibliography}

\end{document}